\documentclass[10pt,twocolumn,letterpaper]{article}

\usepackage{cvpr}              

\definecolor{cvprblue}{rgb}{0.21,0.49,0.74}
\usepackage[pagebackref,breaklinks,colorlinks,allcolors=cvprblue]{hyperref}
\usepackage{graphicx,subcaption, float}
\usepackage{graphicx}
\usepackage{booktabs}
\usepackage{multirow}   
\usepackage{subcaption}
\usepackage{array}      
\usepackage{siunitx}    
\usepackage[table]{xcolor}

\definecolor{tabletop}{HTML}{1A2E40}     
\definecolor{tablerow}{HTML}{F4F7F9}     
\definecolor{ourmodelrow}{HTML}{E8F0F8}

\usepackage{xcolor}
\usepackage[table]{xcolor}
\usepackage{hyperref}
\hypersetup{
    colorlinks=true,
    linkcolor=magenta,  
    citecolor=magenta,
    urlcolor=magenta
}

\usepackage{url}
\usepackage{xcolor}
\definecolor{newcolor}{rgb}{.8,.349,.1}

\usepackage{algorithm}
\usepackage{algpseudocode}
\usepackage{subcaption}
\usepackage{multirow}
\usepackage{xcolor}
\usepackage{graphics}
\usepackage{booktabs}
\usepackage{bbding}

\title{Unifying Semantic Priors and High-Frequency Traces: Enhancing V-JEPA with Mixture-of-Experts for Robust Synthetic Image Forensics}

\author{
\textbf{Simone Teglia, Irene Amerini} \\
Sapienza University of Rome \\
\textit{\{teglia, amerini\}@diag.uniroma1.it}
}

\begin{document}
\twocolumn[{%
\renewcommand\twocolumn[1][]{#1}%
\maketitle
\begin{center}
    \captionsetup{type=figure}
\end{center}%
}]
\begin{abstract}
The unchecked proliferation of manipulated images on social media platforms has increased the spread of misinformation, posing a severe threat to public trust and information integrity. 
Modern deepfake detectors typically rely on Vision Transformers (ViTs) to capture the low-level inconsistencies that characterize fully synthetic or locally tampered images.
However, the global understanding of such foundation models is not enough to discriminate alone between real and fake multimedia content, especially in challenging scenarios where images are compressed or transmitted through social media. 
In this paper we pioneer the application of Joint-Embedding Predictive Architecture (JEPA) models to deepfake detection, taking advantage of the generalized representation of visual reality that such World Models have exhibited. 
We hypothesize, and empirically demonstrate, that the intrinsic world understanding of JEPA models can be used as a strong prior for a deepfake detector. 
To fully exploit JEPA capabilities, we propose MoE-JEPA, a dual-stream architecture for deepfake detection. By enhancing a V-JEPA 2 backbone with a Residual Mixture-of-Experts (MoE) mechanism, along with a noise stream branch, our model dynamically internalizes forensic knowledge. Furthermore, a Gated Attention Multiple Instance Learning (MIL) module is employed to ensure precise spatial semantic understanding.
Evaluated on the SID-Set benchmark, comprising 300K AI-generated, tampered and authentic images, MoE-JEPA establishes a new state-of-the-art with an accuracy of 95.54\%, successfully outperforming vastly larger models. 
\end{abstract}    
\section{Introduction}
\label{sec:intro}
\begin{figure*}[!ht]
    \centering
    \includegraphics[width=0.95\linewidth]{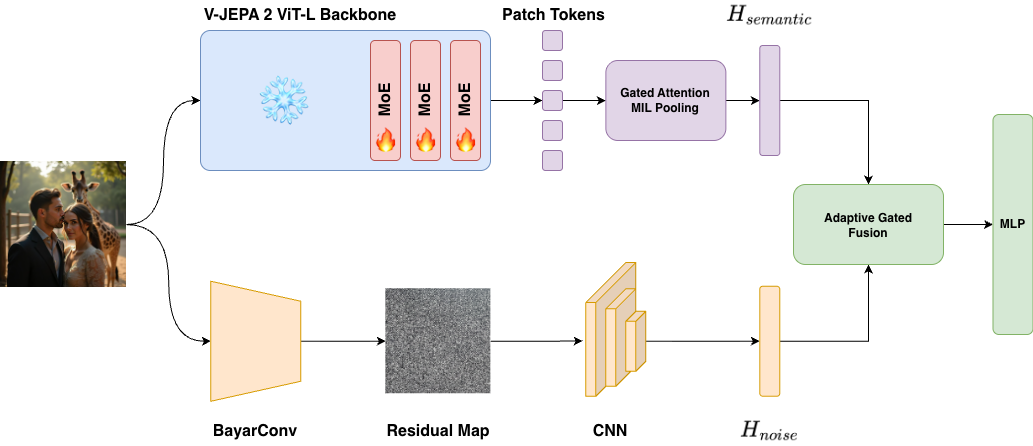}
    \caption{Overview of the MoE-JEPA architecture. The semantic branch (top) uses a V-JEPA 2 backbone, a MoE module, and a Gated Attention MIL module to semantically suppress pristine backgrounds and isolate localized tampering ($H_{\text{semantic}}$). In parallel, the noise branch (bottom) applies a constrained BayarConv layer to extract high-frequency generative and sensor artifacts ($H_{\text{noise}}$). These complementary features are joined using an Adaptive Gated Multimodal Fusion module and classified via an MLP head.}
    \label{fig:architecture}
\end{figure*}

With the rapid advancements of Generative Artificial Intelligence, the creation of artificial multimedia content has witnessed explosive growth. State-of-the-art models like DALL-E 3 \cite{betker2023improving}, FLUX \cite{labs2025flux} and Nano Banana Pro \cite{team2023gemini}, have democratized the access to such powerful tools, making the generation of highly realistic images, commonly known as deepfakes, no longer a research novelty, but an everyday technology.
However, the same technological progress that enables creative expression and productivity gains, also introduces profound societal risks. Alongside these technological breakthroughs, society has witnessed an unprecedented wave of digital disinformation propagating across social media platforms \cite{jimaging11030073, 10.1145/3589334.3649116, explodingdeepfakes}. Synthetic multimedia content is regularly weaponized to manipulate public perception, posing a severe threat to public trust, democratic processes, and overall information integrity. 
Fabricated images depicting sensitive geopolitical conflicts, such as those in Ukraine or Gaza, can rapidly go viral on social media, potentially altering narrative and concretely influencing people's beliefs, driving real-world actions \cite{10.1371/journal.pone.0320124}. 
These concerns have intensified research efforts in multimedia forensics, prompting the release of massive deepfake datasets and the development of increasingly powerful detection architectures.
Modern detectors, often built upon large-scale vision backbones, aim to discern authentic images from AI-generated or manipulated ones by learning discriminative artifacts introduced during the generation process. In particular, foundation Vision Transformers (ViTs) \cite{dosovitskiy2021an} have emerged as powerful general-purpose feature extractors, achieving remarkable performance across a wide range of visual recognition tasks.
By modeling long-range dependencies through self-attention, ViTs primarily learn global semantic representations of the image. While this property is advantageous for high-level understanding, it may limit their ability to capture fine-grained, localized inconsistencies that are critical in forensic analysis. When applied to the field of deepfake detection, such models can effectively capture global generative patterns that characterize fully synthetic images, but their strong semantic abstraction often suppresses subtle forensic cues that fine-grained detectors rely upon \cite{wang2024timelysurveyvisiontransformer}. 
At the same time, recent progress in self-supervised learning have further improved the transformer-based backbones. 
In particular Joint-Embedding Predictive Architecture (JEPA) \cite{lecun_path} models propose a new learning paradigm that involves the encoder to learn to anticipate the representation of masked regions of the images in the latent space, rather than the actual raw pixels. Architectures of this kind serve as a foundational component for building World Models, as their predictive module aims to capture the underlying semantic structure and constraints of the visual world.
We hypothesize that this learned generalized understanding of images can act as a fundamental prior to capture the synthetic inconsistencies of AI-generated imagery. To harness this capability, we introduce MoE-JEPA (see Figure \ref{fig:architecture}), a dual-stream architecture for deepfake detection based on the World Model V-JEPA 2 \cite{VJEPA2}, a self-supervised vision model that excel at both video and image tasks. Specifically, we employ the V-JEPA 2 backbone to exploit its robust visual representations, and augment its capabilities by injecting into the latest layers a Residual Mixture-of-Experts (MoE) mechanism, allowing the architecture to internalize forensic features the backbone was not trained to extract. Furthermore, to prevent the semantic dilution caused by global pooling that can influence the correct detection of localized forgeries, MoE-JEPA implements a Gated Attention Multiple Instance Learning (MIL) module to ensure precise, unsupervised spatial routing to manipulated regions. 
Furthermore, we extend the architecture with a low-level noise branch, enabling a richer understanding of the image and making the detection more robust, especially against the aggressive degradations introduced by social media sharing, resizing, and transmission. By explicitly capturing localized generative and sensor artifacts, this parallel noise stream compensates for the semantic abstraction of the foundation model, providing a resilient defense against in-the-wild media degradation.

In summary, our main contributions are:

\begin{itemize}
    \item To the best of our knowledge, we are the first to pioneer the usage of JEPA World Models for deepfake detection. We empirically demonstrate that the intrinsic understanding of physical and structural visual reality learned by these models naturally captures generative inconsistencies.

    \item We propose \textbf{MoE-JEPA}, a deepfake detector combining a semantic branch with a noise branch. By integrating a Residual Mixture-of-Experts (MoE) and a Gated Attention Multiple Instance Learning (MIL) module, the architecture dynamically internalizes forensic traces and mitigates background semantic dilution without catastrophic forgetting of the foundation model's pre-trained knowledge.

    \item We design a robust dual-stream framework with an Adaptive Gated Fusion technique. We demonstrate that while V-JEPA 2 excel at global contextual verification, coupling them with a constrained BayarConv noise branch is critical for isolating semantically coherent, microscopic local manipulations.

    \item MoE-JEPA achieves a new state-of-the-art on the SID-Set benchmark, with a \textbf{95.54\%} overall accuracy. Tested on the RRDataset, it achieves the second best accuracy overall and the best accuracy on re-digitalized real images.

\end{itemize}

\section{Related Work}

\subsection{Deepfake Detection}
In recent years, the emergence of powerful generative artificial intelligence models like GANs \cite{gan-goodfellow2014} and Diffusion Models \cite{NEURIPS2020_4c5bcfec} has called for the development of reliable and effective deepfake detectors. Early deepfake detection methods primarily relied on Convolutional Neural Networks (CNNs) to expose spatial artifacts, blending boundaries, and frequency-domain inconsistencies. \cite{afchar2018mesonet, liu2020global, zhong2023patchcraft, tan2023learning}
As generative models advanced, researcher started to adopt foundation Vision Transformers (ViTs) \cite{dosovitskiy2021an} as base model for their architectures \cite{dong2022protecting, 10.1007/978-3-031-20065-6_23,10.1145/3512527.3531415, electronics11244143}. Latest approaches leverage more complex architectures to extract image embeddings, like Large Vision Language Models \cite{huang2025sida} or Q-Former \cite{li2023blip}. These approaches take advantage of the global context understanding to achieve superior face forgery detection performance. However, these large models introduce prohibitive computational costs, making real-time inference for social media filtering impractical. To overcome these deployment hurdles we explore the usage of the Joint-Embedding Predictive Architecture as vision encoder, comprising only of 0.3B parameters, providing a remarkable lightweight and efficient foundation model.

\subsection{Mixture of Experts}
Mixture-of-Experts (MoE) \cite{6797059} aim to increase the model capacity without affecting the computational demand, by dynamically routing input tokens to a specialized subset of feed-forward sub-networks (experts). To drastically scale up the model capacity, without affecting the inference costs, Shazeer et al. \cite{shazeer2017outrageously} proposed the sparsely-gated Mixture-of-Experts, that select only the \textit{top-k} most relevant experts for each given token. 
This allows the overall parameter count of the architecture to scale significantly, while the computational complexity remains equivalent to a much smaller dense model. 
Moreover, to protect the foundation model's universal knowledge, the Residual Mixture-of-Experts \cite{wu2022residualmixtureexperts} paradigm introduced a set of frozen shared experts, responsible to maintain the pretrained knowledge of the model, alongside a dynamically routed trainable experts. This allows for a more stable training, enabling the specialization of routed experts without losing the core semantic understanding of the model.
While traditional MoEs utilize a Softmax routing function, recent advancements like DeepSeekMoE \cite{nguyen2025deepseekmoe} have demonstrated that Normalized Sigmoid Gating prevents the possible routing collapse.
In the field of multimedia forensics, works like MoE-FFD \cite{kong2025moe} and Forensic-MoE \cite{fang2025forensic} have successfully utilized expert modules to extract generalized face forgery clues and aggregate diverse synthetic traces from multiple generators. 

\subsection{Joint-Embedding Predictive Architectures}
The Joint-Embedding Predictive Architecture (JEPA), originally proposed by LeCun as a foundational step toward Autonomous Machine Intelligence \cite{lecun_path}, represents a paradigm shift in self-supervised representation learning world. Unlike traditional generative approaches which attempt to reconstruct missing patches at the exact pixel or token level, JEPA models operate entirely within an abstract latent space, by predicting the representation of missing parts of the inputs. 
Extending the base capabilities of JEPA models, I-JEPA \cite{IJEPA} and V-JEPA \cite{VJEPA} demonstrated that this abstract prediction strategy yields highly scalable and semantically rich representations even for images and videos respectively, by predicting latent visual features rather than raw pixels. 
Building upon this foundation, V-JEPA 2 \cite{VJEPA2} extends the V-JEPA framework, successfully obtaining a new state-of-the-art for vision JEPA models. Because JEPA models are explicitly trained to understand the underlying physical rules, motion dynamics, and structural consistencies of visual reality, we hypothesize that their latent spaces inherently capture the anomalous and physically inconsistent nature of synthetic media.

\section{Methodology}
\label{sec:methodology}
We propose MoE-JEPA, a deepfake detection architecture that leverages the natural vision understanding of the World Model V-JEPA 2. By combining a semantic branch, devoted to extracting high-level structural anomalies and profound physical inconsistencies from the latent space, and a noise branch, responsible for isolating microscopic, high-frequency generative traces, our architecture is able to evaluate the image in its entirety, comprehending both high and low-level information at the same time. 

\subsection{The Semantic Branch}
\subsubsection{Pre-trained Vision Backbone}
Given an input image $I \in \mathbb{R}^{H \times W \times C}$, we first adapt the input for the temporal requirement of V-JEPA by duplicating the frame to simulate a video stream across a synthetic temporal dimension $T$. Then we process it through a frozen V-JEPA Encoder model $E$, which outputs a sequence of unpooled spatial patch tokens: 
$$Z = E(I) \in \mathbb{R}^{N \times D}$$
where $N$ is the number of spatial patches and $D$ is the embedding dimension. The first layers of the Vision Encoder $E$ are kept frozen to prevent catastrophic forgetting of general visual representations and to preserve its large-scale pretraining knowledge. 

\subsubsection{Residual Mixture-of-Experts}
Rather than completely relying on the generalized visual understanding of the pre-trained backbone, which is inherently optimized for semantic invariance rather than forensic sensitivity, we intervene in the latest transformer blocks to increase adaptability to forensic cues. Specifically we replace the last Multi Layer Perceptron (MLP) of the latest encoder blocks of the vision backbone with a Residual Mixture-of-Experts (MoE) module. Concretely, each MoE layer consists of a set of $M$ trainable experts $\{E_1, E_2, \dots, E_M\}$, together with a single fully frozen shared expert $E_{\text{shared}}$, following a Residual Mixture-of-Experts formulation. The shared expert remains always active and provides a stable semantic pathway, while the trainable experts are dynamically routed to model input-specific variations. To control the routing dynamic we adopt a Normalized Sigmoid Gating mechanism, following recent studies \cite{nguyen2025deepseekmoe} that prove that this kind of routing allows for a more uniform saturation across layers than the softmax-gated alternative, promoting a more balanced expert utilization and faster router stabilization.

Formally, let $z_i \in \mathbb{R}^D$ denote a spatial patch token. A learnable router network parameterized by a weight matrix $W_R \in \mathbb{R}^{D \times M}$ first computes raw affinity logits for all trainable experts $s(z_i) = W_R z_i$.
Let then $\mathcal{T}_k(z_i)$ denote the selected subset of the Top-$k$ most relevant experts, and let $g_m(z_i)$ represent their corresponding normalized sigmoid gating weights. The final adapted token $z_i^{\text{out}}$ is computed as the sum of the deterministic shared expert representation and the dynamically routed trainable experts:$$z_i^{\text{out}} = E_{\text{shared}}(z_i) + \sum_{m \in \mathcal{T}_k(z_i)} g_m(z_i) \cdot E_m(z_i)$$

To encourage diversity among experts and avoid degenerate specialization, we perturb the weights of the trainable experts at initialization by injecting small stochastic noise. This controlled perturbation promotes heterogeneous adaptation trajectories, enabling the model to better capture diverse input distributions. To prevent routing collapse and ensure that all the trainable experts are equally utilized, we employ an auxiliary load balancing loss $\mathcal{L}_{bal}$. Given a batch of $B$ tokens and $M$ trainable experts, we first compute the routing density $d_{m}$ for each expert $m$ as the mean routing probability across all tokens:$$d_{m} = \frac{1}{B} \sum_{i=1}^{B} p_{i,m}$$where $p_{i,m}$ represents the routing probability of token $i$ to expert $m$. The load balancing loss is then computed as the scaled sum of the squared densities:$$\mathcal{L}_{bal} = M \sum_{m=1}^{M} d_{m}^{2} - 1$$
The total optimization objective of the network is thus $\mathcal{L}_{total} = \mathcal{L}_{CE} + \lambda_{bal}\mathcal{L}_{bal}$, where $\mathcal{L}_{CE}$ is the primary Cross-Entropy loss and $\lambda_{bal}$ is the balancing coefficient.

\subsubsection{Gated Attention MIL Pooling}
Finally, to isolate the tampered regions and mitigate semantic dilution, we apply a Gated Attention Multiple Instance Learning (MIL) module over the adapted tokens. This module acts as a spatial circuit breaker, dynamically learning to assign near-zero attention weights to pristine background patches, making the model attend more easily to synthetically forged patches. 
The attention score $a_i$ for each adapted patch token $z_i^{\text{out}}$ is computed via a dual-branch network:
$$a_i = \text{Softmax}\left( w^T \left[ \tanh(V z_i^{\text{out}}) \odot \sigma(U z_i^{\text{out}}) \right] \right)$$
with $V$ and $U$ as learnable projection matrices that map the token into a forensic feature space and a relevance gate, respectively. The $\tanh$ activation models complex feature interactions, while the Sigmoid gate $\sigma$ explicitly masks out irrelevant semantic content via element-wise multiplication ($\odot$). The final localized semantic representation $H_{semantic}$ is the attention-weighted sum of the tokens:$$H_{\text{semantic}} = \sum_{i=1}^{N} a_i z_i^{\text{out}}$$

\subsection{The Noise Branch}
While the semantic branch demonstrates strong capabilities in isolating spatial and contextual anomalies, relying exclusively on high-level semantic features can limit the model robustness. For example, many sophisticated forgeries are designed to be semantically coherent: manipulated regions may exhibit perfectly blended boundaries, consistent lighting, and plausible object structure. Moreover, common post-processing steps, such as resizing, recompression, and platform-specific social media pipelines, can attenuate the semantic inconsistencies that detectors often rely on. To enhance the overall architecture robustness and enable a more comprehensive understanding of the forgeries, we introduce a parallel noise branch. Rather than processing the latent patch tokens, this branch operates directly on the raw input image $I \in \mathbb{R}^{H \times W \times C}$ to extract a purely high-frequency residual map. 

To extract the noise residuals we employ a constrained convolutional layer, commonly known as BayarConv \cite{bayar2016deep}. This layer exploits a learnable convolutional kernel $W \in \mathbb{R}^{K \times K}$ that is forced to predict the center pixel value, based only on its neighborhood pixels. This suppresses dominant semantic structures, effectively filtering out the image semantic and revealing the subtle high-frequency noise patterns beneath.
The resulting residual map $R$ is subsequently processed through a lightweight stack of standard convolutional blocks. These blocks serve to hierarchically pool the localized high-frequency discrepancies into a dense, abstract frequency fingerprint $H_{\text{noise}} \in \mathbb{R}^{d}$.

By fusing this frequency-based fingerprint with the spatially localized semantic representation, the architecture achieves a comprehensive, disentangled view of the image: the semantic branch evaluates contextual integrity and semantic coherence, while the noise branch provides a more robust verification of physical and generative consistency. Together, they provide a more comprehensive understanding of the image, improving the detection of both fully synthetic images and carefully crafted local manipulations.

\subsubsection{Adaptive Gated Multimodal Fusion and Classification}
Rather than just relying on a simple concatenation strategy, we employ an Adaptive Gated Residual Fusion mechanism. 
After processing the image $I$ through the semantic branch and the noise branch, to dynamically regulate the contribution of the  the two vectors $H_{\text{semantic}} \in \mathbb{R}^D$ and $H_{\text{noise}} \in \mathbb{R}^d$, we compute a modulating gate vector $\alpha$. This is achieved by concatenating the normalized representations and passing them through a learnable dense projection followed by a Sigmoid activation:$$\alpha = \sigma\left(W_{\text{gate}} \left[ \bar{H}_{\text{semantic}} \oplus \bar{H}_{\text{noise}} \right] \right)$$where $\oplus$ denotes concatenation.

Finally the multimodal representation is constructed by applying the learned gate $\alpha$ via element-wise multiplication ($\odot$) to the noise embedding, which is then added to the semantic embedding:

$$H_{\text{fused}} = \bar{H}_{\text{semantic}} + (\alpha \odot \bar{H}_{\text{noise}})$$

This gated formulation allows the network to autonomously decide when to rely purely on spatial anomalies and when to inject microscopic generative noise. Later $H_{\text{fused}}$ is fed to a linear classifier to predict the target classes. 
\section{Experimental Setup}
\label{sec:experimental}

\subsection{Dataset}
To fully evaluate the discriminative capabilities and cross-domain robustness of the proposed architecture we conduct extensive experiments across two different benchmarks. The following datasets have been chosen due to their complementary nature and their ability to extensively reflect the modern landscape of manipulated content shared across social media platforms. SID-Set offers a comprehensive collection of contemporary AI-generated and tampered media, required to required to test baseline detection accuracy, while RRDataset focuses on severe image degradation techniques, allowing to evaluate model resilience against real-world conditions.

\begin{figure*}[!ht]
\centering
    \begin{subfigure}[b]{0.32\linewidth}
        \centering
        \includegraphics[width=\linewidth]{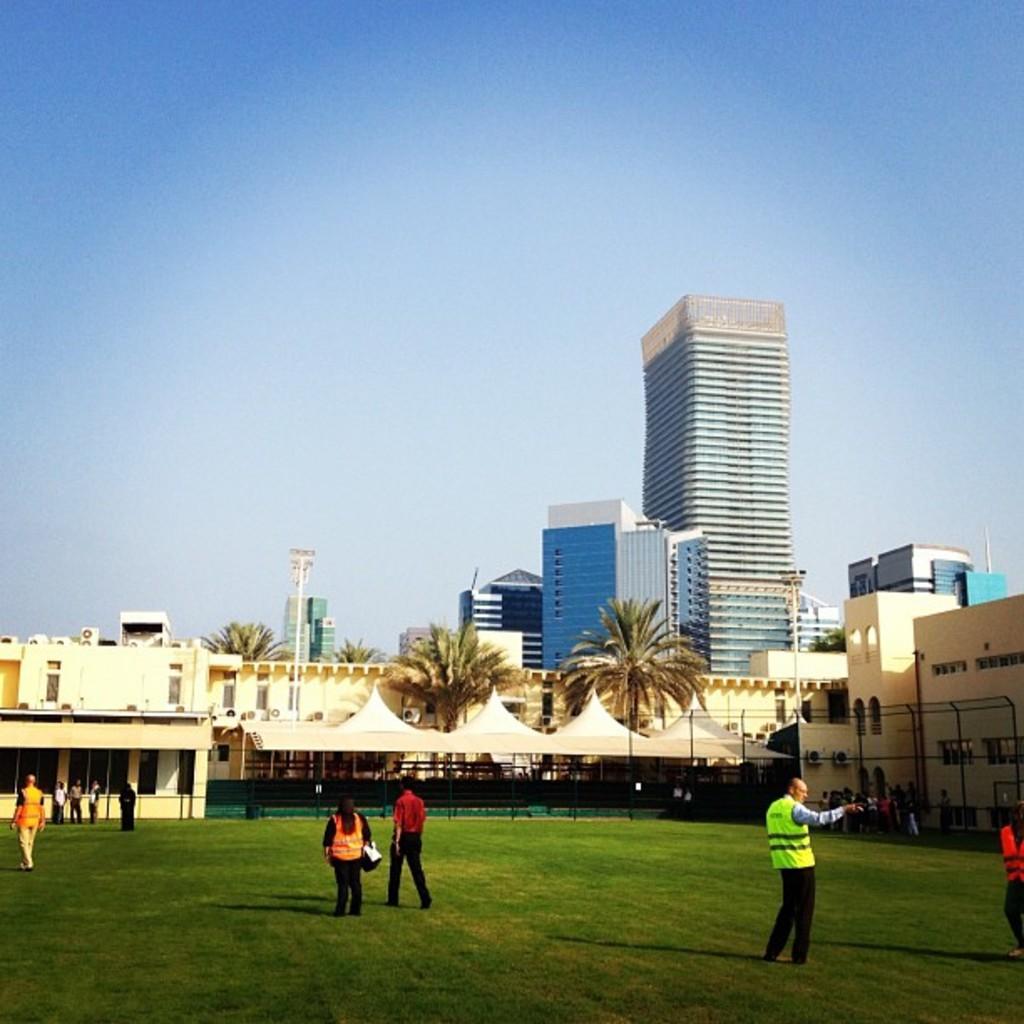}
        \caption{SID-Set - Real}
        \label{fig:row1_img1_orig}
    \end{subfigure}
    \hfill 
    \begin{subfigure}[b]{0.32\linewidth}
        \centering
        \includegraphics[width=\linewidth]{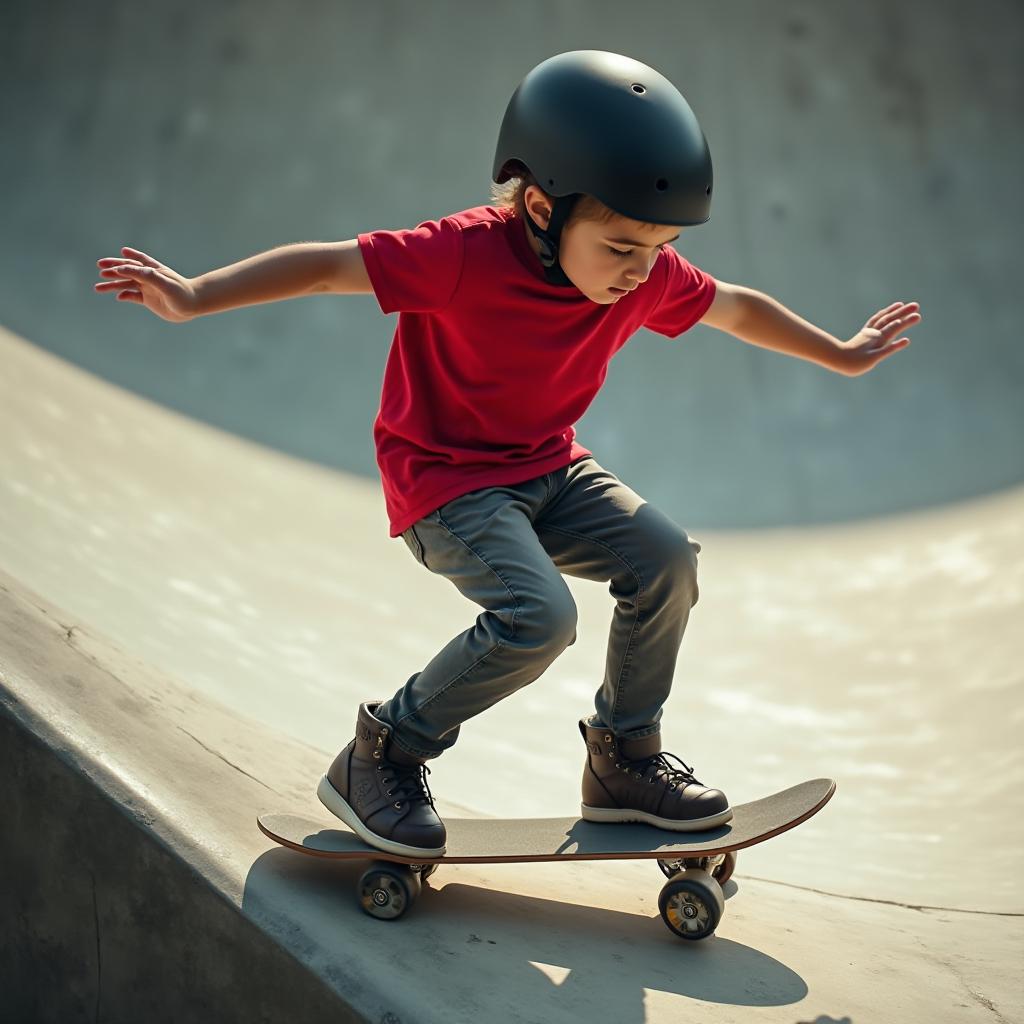}
        \caption{SID-Set - Fully Synthetic}
        \label{fig:row1_img2_trans}
    \end{subfigure}
    \hfill
    \begin{subfigure}[b]{0.32\linewidth}
        \centering
        \includegraphics[width=\linewidth]{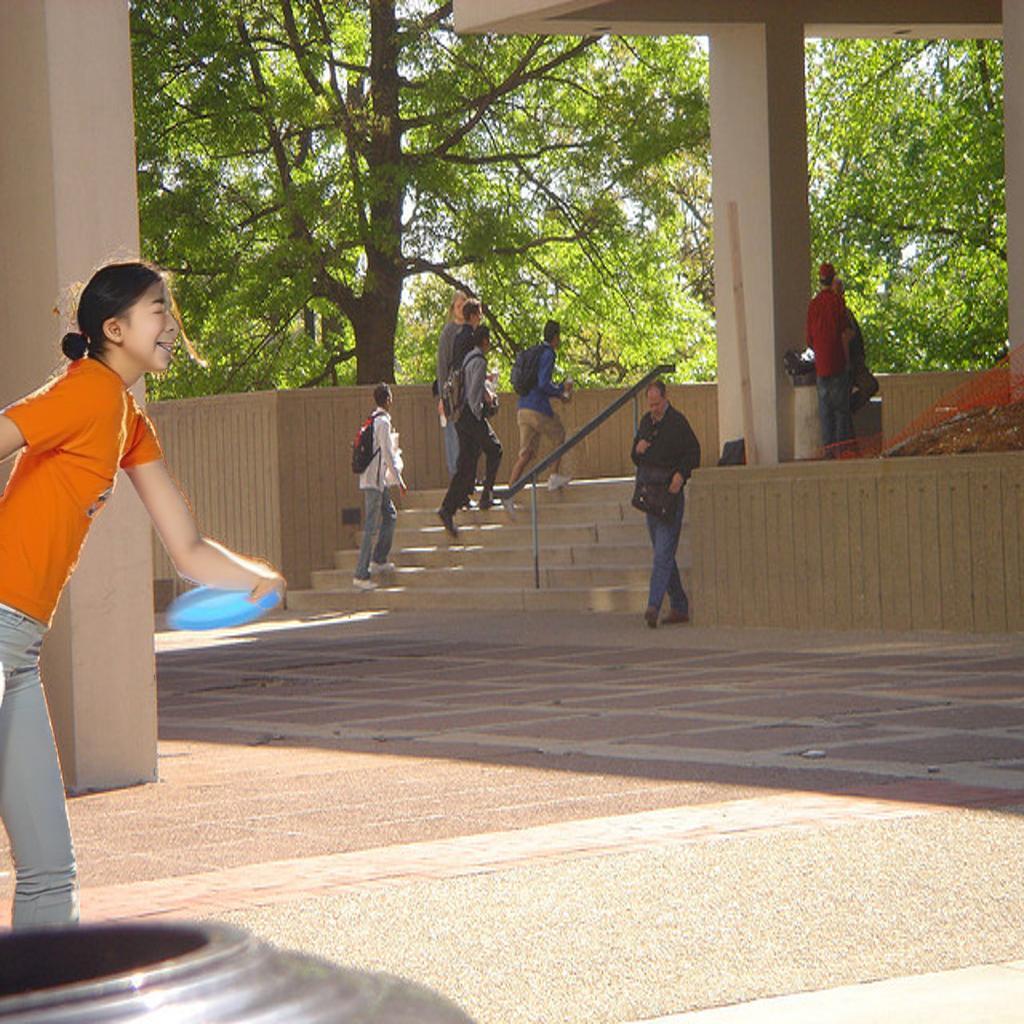}
        \caption{SID-Set - Tampered}
        \label{fig:row1_img3_redig}
    \end{subfigure}
    
    \vspace{0.2cm} 
    
    \begin{subfigure}[b]{0.32\linewidth}
        \centering
        \includegraphics[width=\linewidth]{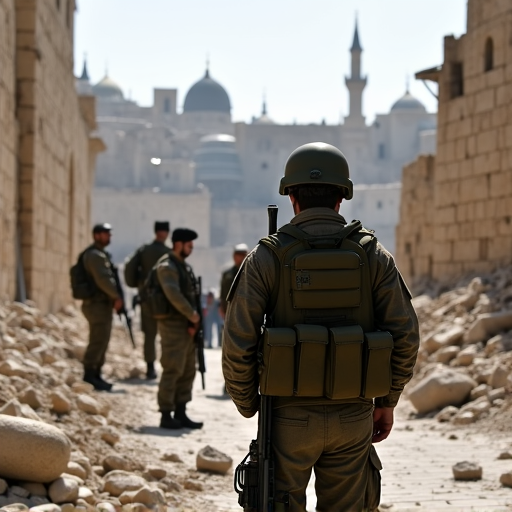}
        \caption{RRDataset - Original}
        \label{fig:row2_img1_orig}
    \end{subfigure}
    \hfill
    \begin{subfigure}[b]{0.32\linewidth}
        \centering
        \includegraphics[width=\linewidth]{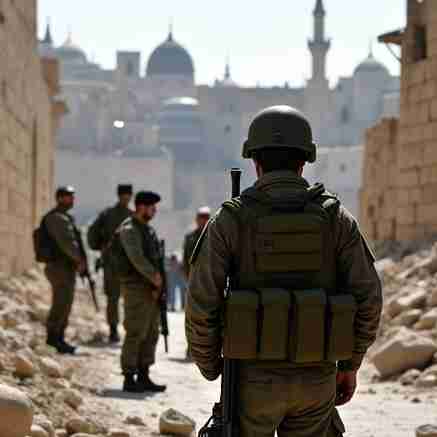}
        \caption{RRDataset - Transmission}
        \label{fig:row2_img2_trans}
    \end{subfigure}
    \hfill
    \begin{subfigure}[b]{0.32\linewidth}
        \centering
        \includegraphics[width=\linewidth]{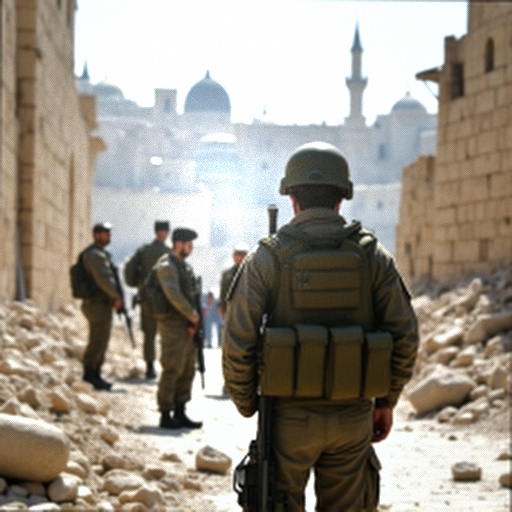}
        \caption{RRDataset - Re-digital}
        \label{fig:row2_img3_redig}
    \end{subfigure}

\caption{Visual examples from the SIDSet and RRDataset. The top row presents images from the SID-Set benchmark across the three categories: \textit{Real, Fully Synthetic} and \textit{Tampered}, while the bottom row displays an image of RRDataset in the three degradation split present in the benchmark: the pristine \textit{Original} format, the \textit{Transmission} split (subject to social media compression), and the \textit{Re-Digitalization} split (captured via screen recapture).}
\label{fig:rrdataset_splits}
\end{figure*}

\subsubsection{SID-Set} For our primary evaluation we employ the Social Media Image Detection Dataset (SID-Set) \cite{huang2025sida}. SID-Set is specifically designed to mimic the complexity of the modern social media landscape, containing highly realistic images uniformly distributed across three categories: Authentic (\textit{Real}), Fully Synthetic (\textit{Fake}), and Locally Manipulated (\textit{Tampered}), making it one of the most comprehensive and up to date deepfake benchmark.
Crucially, because its content is sourced directly from various social media platforms, the dataset inherently captures the real-world degradations such as aggressive algorithmic compression, dynamic resizing, and multiple transmission artifacts that deepfake detectors must overcome "in-the-wild."
We employ the official dataset splits, utilizing 210k images for the training phase, 30k for validation, and 60k for testing.

\subsubsection{RRDataset} To further asses the forensics capabilities of our framework against real-world degradation and unseen forensic cues, we evaluate MoE-JEPA using the RRDataset. This dataset is explicitly designed to test detector resilience against the aggressive post-processing procedures commonly encountered in social-media environment. RRDataset is categorized into three subsets of images: the \textit{Original} split containing pristine images, the \textit{Transmission} split containing images that have been shared through popular messaging and social-media platforms for multiple rounds, and \textit{Re-Digitalization} split comprising of images that have been physically recaptured from screens. 
Moreover, RRDataset comprehends different levels of image sensitivity, involving war, political, religious and disaster scenarios. Testing across these three specific distributions allows us to rigorously measure the capacity of the architecture to survive severe artifact attenuation and complex domain shifts.

\subsection{Implementation Details}
\paragraph{Architecture Configuration: } We instantiate MoE-JEPA using the large-scale V-JEPA 2 foundation model, specifically the \texttt{facebook/vjepa2-vitl-fpc64-256} checkpoint \footnote{\url{https://huggingface.co/facebook/vjepa2-vitl-fpc64-256}}. To introduce forensic adaptability, we inject Residual MoE modules exclusively into the final 3 transformer blocks. Each MoE layer is configured with 1 frozen shared expert and $M = 6$ trainable experts. The router utilizes a Top-$k$ selection strategy with $k = 2$, meaning each token is processed by the shared expert and the 2 most relevant trainable experts. While the entire architecture consists of 480.46M parameters, the Residual MoE layers allow the model to only activate 379.73M parameters during each forward pass, leaving over 100M parameters inactive per token. This allows MoE-JEPA to comfortably process media on standard consumer hardware, and enables it to outperform massive Vision-Language Models while utilizing a fraction of their active inference parameters.

\paragraph{Training Setup: }The network is optimized using the AdamW optimizer, along with a learning rate scheduler featuring 10\% linear warmup phase. The joint optimization objective dynamically balances the primary Cross-Entropy loss with the MoE Load Balancing auxiliary loss scaled by a coefficient of $\lambda_{bal} = 0.01$. All experiments were conducted on a single NVIDIA RTX 5090 GPU. The architecture was trained for a maximum of 20 epochs. To prevent overfitting and ensure optimal generalization, we implemented an early stopping callback with a patience of 3 epochs, strictly monitoring the validation accuracy to determine the best model checkpoint.

\section{Results}
\label{sec:results}
In this section, we present a comprehensive evaluation of MoE-JEPA, benchmarking its performance against state-of-the-art deepfake detection architectures across diverse forensic scenarios.
We first evaluate MoE-JEPA on the SID-Set benchmark, specifically testing its ability on detecting pristine, fully synthetic and tampered images. 

\begin{table*}[!htbp]
\centering
\caption{Detailed Accuracy and F1-Score breakdown across \textbf{SID-Set} categories: Real, Fully Synthetic and Tampered. MoE-JEPA achieves a new state-of-the-art surpassing all the previous approaches tested on the benchmark. Best results are highlighted in \textbf{bold}, and second best are \underline{underlined}.}
\label{tab:sid_set_f1_total_breakdown}
\footnotesize 
\setlength{\tabcolsep}{8pt} 

\begin{tabular}{l cc cc cc cc}
\toprule
\multirow{2}{*}{\textbf{Model}} & \multicolumn{2}{c}{\textbf{Real}} & \multicolumn{2}{c}{\textbf{Fully Synth.}} & \multicolumn{2}{c}{\textbf{Tampered}} & \multicolumn{2}{c}{\textbf{Total}} \\
\cmidrule(lr){2-3} \cmidrule(lr){4-5} \cmidrule(lr){6-7} \cmidrule(lr){8-9}
 & Acc. & F1 & Acc. & F1 & Acc. & F1 & Acc. & F1 \\
\midrule
AntifakePrompt \cite{chang2023antifakeprompt} & 88.90 & 89.10 & 97.50 & 97.90 & 90.90 & 80.40 & 92.40 & 89.10 \\
CNNSpot \cite{wang2020cnn} & 89.00 & 90.80 & 90.70 & 88.10 & 68.10 & 64.00 & 82.60 & 85.40 \\
FreDect \cite{frank2020leveraging} & 46.00 & 47.60 & 60.90 & 66.00 & 37.10 & 53.00 & 47.60 & 55.40 \\
Fusing \cite{ju2022fusing} & 89.20 & \underline{92.07} & 88.10 & 89.10 & 27.00 & 31.40 & 68.10 & 71.00 \\
Gram-Net \cite{liu2020global} & 89.20 & 91.70 & 97.90 & 98.60 & 89.90 & 86.90 & 92.10 & 92.40 \\
UnivFD \cite{ojha2023towards} & 68.03 & 68.50 & 86.40 & 98.00 & 92.50 & 90.00 & 82.40 & 85.40 \\
LGrad \cite{tan2023learning} & 62.00 & 76.10 & 58.00 & 67.30 & \textbf{98.90} & \textbf{98.80} & 73.00 & 80.60 \\
LNP \cite{bi2023detecting} & 14.40 & 23.00 & 36.20 & 35.60 & 93.30 & \underline{94.60} & 48.00 & 51.10 \\
SIDA-7B \cite{huang2025sida} & 89.10 & 91.00 & \underline{98.70} & 98.60 & 92.70 & 91.00 & 93.50 & \underline{93.50} \\
SIDA-13B \cite{huang2025sida} & \underline{89.60} & 91.10 & 98.50 & \underline{98.70} & 92.90 & 91.20 & \underline{93.60} & \underline{93.50} \\
\midrule
\rowcolor{ourmodelrow}
\textbf{MoE-JEPA (Ours)} & \textbf{91.61} & \textbf{93.06} & \textbf{99.98} & \textbf{99.67} & \underline{95.12} & 93.50 & \textbf{95.54} & \textbf{94.21} \\
\bottomrule
\end{tabular}
\end{table*}
As clearly visible in Table \ref{tab:sid_set_f1_total_breakdown}, our proposed architecture achieves a new state-of-the-art performance, with an overall accuracy of 95.54\% and F1-score of 94.21\%, outperforming greatly larger models such as SIDA-13B. Crucially, MoE-JEPA exhibits unparalleled balance across the three different categories. While highly specialized models like LGrad \cite{tan2023learning} and Gram-Net \cite{liu2020global} achieve marginally higher accuracy on the Tampered class (98.90\% and 98.60\%, respectively), they suffer from severe mode collapse, failing catastrophically on the Real and Fake distributions. This indicates a high sensitivity to local noise, making them not robust for real-world environment deployment. Conversely, our dual-stream architecture is able to extract semantics and noise, making it highly accurate on all types of images. This proves that the Gated Attention MIL and the MoE router effectively prevent the semantic dilution that plagues traditional global pooling methods.

\begin{table*}[htbp]
\centering
\caption{Detailed Performance Comparison on the \textbf{RRDataset} Categories. All metrics are reported as Accuracy (\%). Among the detectors, best results are highlighted in \textbf{bold}, and second best are \underline{underlined} }
\label{tab:rrdataset_accuracy}
\scriptsize
\setlength{\tabcolsep}{8pt} 
\renewcommand{\arraystretch}{1.2}   

\begin{tabular}{l *{7}{S}}
\toprule
\multirow{3}{*}{\textbf{Model}} & \multicolumn{7}{c}{\textbf{RRDataset}} \\
\cmidrule(lr){2-8}
 & \multicolumn{2}{c}{\textbf{Original}} & \multicolumn{2}{c}{\textbf{Transmission}} & \multicolumn{2}{c}{\textbf{Re-Digitalization}} & {\textbf{Overall}} \\
\cmidrule(lr){2-3} \cmidrule(lr){4-5} \cmidrule(lr){6-7}
 & {\textit{Real}} & {\textit{Fake}} & {\textit{Real}} & {\textit{Fake}} & {\textit{Real}} & {\textit{Fake}} & \\
\midrule
\rowcolor{tablerow}
\multicolumn{8}{l}{\textit{Vision-Language Models (Evaluated in a Zero-Shot configuration)}} \\
\midrule
Grok-2-vision \cite{grok2} & 46.15 & 91.84 &  52.12 & 94.03 & 48.01 & 81.63 & 69.96 \\
Gemini-2-flash \cite{team2023gemini} & 71.10 & 98.43 & 52.19 & 97.41 & 46.11 & 97.41 & 77.28 \\
Claude-3.7-sonnet \cite{claude} & 85.12 & 94.57 & 71.26 & 96.17 & 62.34 & 85.41 & 82.48 \\
GPT-4o-latest \cite{achiam2023gpt} & 96.30 & 92.68 & 79.41 & 90.01 & 69.23 & 76.92 & 84.09 \\
\midrule
\rowcolor{tablerow}
\multicolumn{8}{l}{\textit{Detectors (Train on GenImage-SDv1.4 \& fine-tune on RRDataset)}} \\
\midrule
LGrad \cite{tan2023learning} & 51.00 & 81.29 & 18.86 & 92.54 & 14.71 & 88.27 & 57.78 \\
UnivFD \cite{ojha2023towards} & 64.79 & 64.90 & 44.61 & 70.80 & 36.15 & 75.69 & 59.49 \\
Fusing \cite{ju2022fusing} & 87.24 & 92.46 & 7.38 & \textbf{99.04} & 30.79 & 73.97 & 65.15 \\
FreDect \cite{frank2020leveraging} & 79.60 & 75.93 & 58.13 & 82.11 & 46.34 & 69.95 & 68.68 \\
LNP \cite{bi2023detecting} & 83.14 & 89.26 & 38.23 & 89.30 & 31.91 & 91.05 & 70.48 \\
CNNSpot \cite{wang2020cnn} & 72.42 & 89.09 & 65.72 & 88.78 & 43.12 & 86.72 & 74.31 \\
Gram-Net \cite{liu2020global} & 81.34 & 74.65 & 79.49 & 75.69 & 79.45 & 62.02 & 75.44 \\
DIRE \cite{wang2023dire} & 89.72 & \textbf{98.25} & \underline{90.34} & \underline{97.87} & 1.42 & \textbf{98.89} & 79.42 \\
DRCT-ConvB \cite{chen2024drct} & \textbf{93.52} & \underline{95.52} & \textbf{92.82} & 95.09 & \underline{64.34} & \underline{96.22} & \textbf{89.59} \\
\midrule
\rowcolor{ourmodelrow}
\textbf{MoE-JEPA (Ours)} & \underline{91.36} & 94.51 & 81.16 & 84.18 & \textbf{91.55} & 61.89 & \underline{84.11} \\
\bottomrule
\end{tabular}
\end{table*}
To assess the architecture ability in even more challenging real-world scenarios, we evaluate MoE-JEPA on the RRDataset benchmark. Following the training protocol proposed by the authors, we pre-train the architecture using 162,000 images from GenImage-SDv1.4 \cite{zhu2023genimage} and 162,000 images from ImageNet. Later, we finetune the best checkpoint on the training split of RRDataset. To ensure a rigorous comparison, all the baselines, including the VLMs, are reported from the official RRDataset benchmark. As shown in Table \ref{tab:rrdataset_accuracy}, MoE-JEPA demonstrate incredible stability across the different post-processing splits of RRDataset, achieving an overall accuracy of 84.11\%, outperforming frontier Vision-Language Models like GPT-4o-latest and Claude-3.7.sonnet. This empirical result underscores the importance of specialized forensic adaptation via Residual MoE when applied to smaller models.  

Evaluation against specialized deepfake detectors highlights the challenging nature of the RRDataset, revealing a strongly fragmented landscape with no single architecture dominating across all different splits. DIRE \cite{wang2023dire}, DRCT-ConvB \cite{chen2024drct}, and MoE-JEPA effectively lead the leaderboard, each exhibiting distinct trade-offs. MoE-JEPA secures a highly competitive second position overall, achieving the best accuracy on the Re-Digitalization \textit{Real} split, successfully resisting the physical noise introduced by such alteration thanks to the robust V-JEPA world model initialization. Ultimately, this demonstrates that our dual-stream approach provides a reliable, balanced, and structurally aware verification for localized noise reconstruction.

\subsection{Ablation Studies}
To empirically validate the architectural design choices of MoE-JEPA, we conduct a comprehensive ablation study on the SID-Set benchmark. By systematically enabling and disabling core modules, we isolate the specific performance contributions of the Mixture-of-Experts (MoE) layers, the Gated Attention MIL pooling, and the high-frequency Noise Stream. The results are summarized in Table 3.
We first evaluate a naive baseline using only the intrinsic knowledge of the frozen V-JEPA encoder. This configuration yields an overall accuracy of 80.27\%. Introducing the noise stream directly into this architecture improves the detection of manipulated images by more than +10\% points, underlining the importance of such branch for detecting high-frequency anomalies that the semantic branch alone misses.
\begin{table*}[!ht]
\centering
\caption{Ablation Study on SID-Set. Impact of architectural components on class-specific and overall test accuracy.}
\label{tab:ablation}
\scriptsize 
\setlength{\tabcolsep}{8pt} 
\begin{tabular}{l c c c | c c c c}
\toprule
\textbf{Model Variant} & \textbf{Pooling} & \textbf{MoE} & \textbf{Noise} & \textbf{Real} & \textbf{Fake} & \textbf{Tampered} & \textbf{Acc.} \\
\midrule
Baseline 1 & Avg. Pooling & $\times$ & $\times$ & 71.59 & 96.77 & 72.46 & 80.27 \\
Baseline 2 & Avg. Pooling & $\times$ & $\checkmark$ & 69.90 & 99.25 & 85.47 & 84.87 \\
Variant A1 & Avg. Pooling & $\checkmark$ & $\times$ & 91.16 & 99.83 & 93.83 & 95.09 \\
Variant B & Avg. Pooling & $\checkmark$ & $\checkmark$ & 92.76 & 99.80 & 92.88 & 95.15 \\
MoE-JEPA (Sem.) & Gated MIL & $\checkmark$ & $\times$ & \textbf{93.43} & \textbf{99.90} & 93.17 & 95.50 \\
\textbf{MoE-JEPA (Full)} & Gated MIL & $\checkmark$ & $\checkmark$ & 91.61 & \textbf{99.90} & \textbf{95.12} & \textbf{95.54} \\
\bottomrule
\end{tabular}
\end{table*}
However, the introduction of the MoE mechanism triggers a massive performance improvement, with the overall accuracy that spikes by nearly +15\% points. Most notably, the accuracy on the \textit{Real} class jumps to 91.16\% and \textit{Tampered} detection leaps to 93.83\%. This proves that the Residual MoE successfully allows the network to internalize domain-specific forensic cues without suffering from the catastrophic forgetting of its pre-trained visual priors.
We then replace the standard average pooling with out Gated Attention MIL pooling module. This introduction boosts the performance of the model across all categories, revealing the importance of spatial routing. By attending to only relevant semantic anomalies, the architecture achieves a highly competitive accuracy and sets the maximum score for the \textit{Real} category. 
Finally, while the semantic only branch of MoE-JEPA already excels at global understanding, the addition of the noise branch allow the network to peak on the \textit{Tampered} class, reaching 95.12\%, while simultaneously securing the highest overall accuracy across all variants at 95.54\%.
Although there is a minor trade-off in \textit{Real} image accuracy, it is outweighed by the enhanced sensitivity to local forgeries and the boost in comprehensive detection. This final configuration proves that the parallel noise stream is strictly necessary to reliably isolate microscopic, semantically coherent local manipulations, yielding the most robust and accurate overall detector.
\section{Conclusions and Future Works}
\label{sec:conclusions}

In this paper we presented MoE-JEPA, a dual-stream architecture for deepfake detection that pioneers the usage of the world model V-JEPA as semantic backbone. By integrating a Residual Mixture-of-Expert mechanism and Gated Attention Multiple Instance Learning (MIL) pooling, MoE-JEPA successfully internalizes domain-specific forensic cues without compromising its robust pre-trained visual world model knowledge. Furthermore, the parallel noise stream extracts high-frequency generative and sensor artifacts, improving the detection of localized forgeries. 
Our extensive evaluation on the SID-Set benchmark demonstrates the excellent capability of our architecture to distinguish between real, fully-synthetic and tampered images, establishing a new state-of-the-art with an overall accuracy of 95.54\%. 
The ablation studies systematically confirm the architectural choices, demonstrating the importance of each module of MoE-JEPA and confirming the necessity of both the dynamic expert routing and the dual-stream formulation.
For future works, we plan to enhance the interpretability of MoE-JEPA by extracting spatial information directly from the processed routed patch tokens, providing localization maps for tampered images. 
{
    \small
    \bibliographystyle{ieeenat_fullname}
    \bibliography{main}
}


\end{document}